\documentclass{article} %
\usepackage{iclr2026_conference,times}

\usepackage{amsmath,amsfonts,bm}

\def\eqref#1{equation~\ref{#1}}

\def\1{\bm{1}}

\DeclareMathAlphabet{\mathsfit}{\encodingdefault}{\sfdefault}{m}{sl}
\SetMathAlphabet{\mathsfit}{bold}{\encodingdefault}{\sfdefault}{bx}{n}

\usepackage{hyperref}
\usepackage{url}
\usepackage{xcolor}
\usepackage{algorithm}
\usepackage{algpseudocode}
\usepackage{cleveref}
\usepackage{graphicx}
\usepackage{tabularx}
\usepackage{booktabs}
\usepackage{natbib}
\usepackage{enumitem}

\title{Conjuring Semantic Similarity}

\author{Tian Yu Liu \\
Department of Computer Science\\
University of California, Los Angeles\\
\texttt{tianyu@cs.ucla.edu} \\
\\
\And
Stefano Soatto \\
Department of Computer Science\\
University of California, Los Angeles\\
\texttt{soatto@cs.ucla.edu} \\
}

\iclrfinalcopy %
\begin{document}

\maketitle
\begin{abstract}
The semantic similarity between sample expressions measures the distance between their latent `meaning'.  These meanings are themselves typically represented by textual expressions. We propose a novel approach whereby the semantic similarity among textual expressions is based {\em not} on other expressions they can be rephrased as, but rather based on the {\em imagery} they evoke.
While this is not possible with humans, generative models allow us to easily visualize and compare generated images, or their distribution, evoked by a textual prompt. Therefore, we characterize  the semantic similarity between two textual expressions simply as the distance between image distributions they induce, or `conjure.'  We show that by choosing  the Jeffreys divergence between the reverse-time diffusion stochastic differential equations (SDEs) induced by each textual expression, this can be directly computed via Monte-Carlo sampling. Our method contributes a novel perspective on semantic similarity that not only aligns with human-annotated scores, but also opens up new avenues for the evaluation of text-conditioned generative models while offering better interpretability of their learnt representations.
\end{abstract}

\section{Introduction}
Semantic similarity is about comparing data not directly, but based on their underlying `concepts' or `meanings'. Since meanings are most commonly expressed through natural language, various methods have attempted to compute them in this space. Words have been often compared based on the occurrences of other words that surround them, and images have likewise been compared `semantically' by using the text captions that describe them.

While measuring semantic similarity comes natural to humans who share significant knowledge and experience, defining semantic similarity for trained models is non-trivial. For the class of Large Language Models (LLMs), \cite{liu2023meaning} defined the space of meanings for autoregressive models as the distribution over model continuations for any given input sequence. Semantic alignment of these models with humans can be directly quantified by comparing the geometric and algebraic properties of this space - for instance, measuring whether prompts semantically similar to humans are also closely embedded in this space of meanings.

In this work, we study the converse: Interpreting the learnt semantic space of image (rather than text) generation models, and how well they align with that of human annotators. Instead of comparing images by the captions that describe them, we compare textual expressions in terms of the images they conjure. In other words, we propose an expanded notion of meaning that is purely ``visually-grounded". This would be hard if not impossible for humans, since the process requires visualizing and comparing  `mental images' each individual can conceive, but it is straightforward for trained models. 

We focus on the class of text-conditioned diffusion models. Under our definition, the semantic similarity between two text passages for a model can simply be measured by the similarity of image distributions generated by the model, conditioned on those passages. 

There is a technical nugget that needs to be developed for the method to be viable, which is how to compare diffusions in the space of images. To address this, we propose to leverage the Jeffreys Divergence between the stochastic differential equations (SDEs) that govern the flow of the diffusion model, which we will show to be computable using a Monte-Carlo sampling approach.

Although our goal is primarily to introduce a method to quantify and visualize the semantic representations learnt by image generation models, we will show that our simple choice of distance already leads to results comparable to zero-shot approaches based on large language models. To validate our proposed definitions, we further conducted ablation studies on several components of our method, demonstrating robustness to the specific decoding algorithms used for multiple choices of diffusion models.

\begin{figure}[t]
    \centering
    \includegraphics[width=0.9\linewidth]{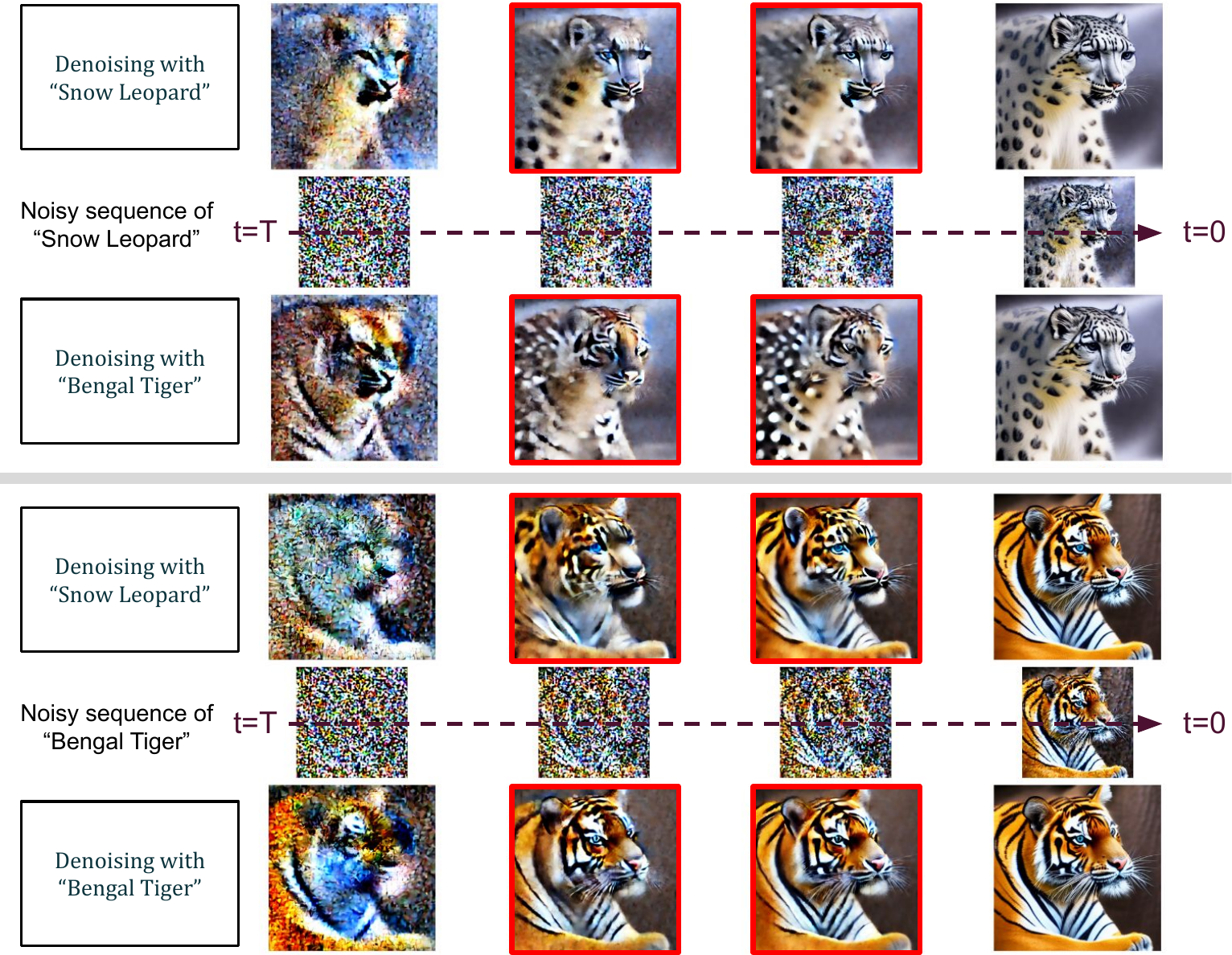}
    \caption{We illustrate the process of conjuring semantic similarity between textual expressions ``Snow Leopard" and ``Bengal Tiger". We denoise each sequence of noisy images (middle row of both halves of figure) with both prompts (top and bottom row of both halves of figure). Our method can be interpreted as taking the Euclidean distance between the resulting images in the two rows. The sequences of noisy images are obtained with either of the two text expressions (top / bottom halves of Figure) starting from a Gaussian prior ($t=T$). Observing cells highlighted in red, we see that the model converts pictures of Snow Leopards into Bengal Tigers by changing their characteristic spotted coats into stripes, and adding striped textures to the animal's face (top half of Figure), and conversely converts Bengal Tigers into Snow Leopards by changing their characteristic stripes into spotted coats (bottom half of Figure). This enables interpretability of their semantic differences via changes in their evoked imageries. }
    \label{fig:conjuring-semantic-similarity}
\end{figure}

To summarize the contributions of this work, we propose an approach for evaluating semantic similarity between text expressions that is grounded in the space of visual images. Our method has a unique advantage over traditional language-based methods that, in addition to providing a numerical score, it also provides a visual expression, or `explanation', for comparison, enabling better interpretability of the learnt representations (\Cref{fig:conjuring-semantic-similarity}). Additionally, our method is the first to enable quantifying the alignment of semantic representations learnt by diffusion models compared to that of humans, which can open up new avenues for the evaluation text-conditioned diffusion models.

\section{Related Works}
\paragraph{Text-Conditioned Image Generative Models.}
We begin by briefly surveying the literature on text-conditioned image generation.
\cite{goodfellow2014generative} proposed the Generative Adversarial Network (GAN), a deep learning-based approach for image generation.
\cite{mirza2014conditional} proposed a method for conditioning GANs based on specified labels. Works such as VQ-VAE \citep{van2017neural} and VQ-VAE-2 \citep{razavi2019generating} also built upon the foundational works of Variational Auto Encoders (VAEs) \citep{kingma2013auto} to learn discrete representations used to generate high quality images, among other outputs, when paired with an autoregressive prior. DALL-E \citep{ramesh2021zero} has also been developed as an effective text-to-image generator leveraging autoregressive Transformers \citep{vaswani2017attention}.

Our work focuses on diffusion models \citep{sohl2015deep}, which have achieved state-of-the-art results among modern image generation models. These models can be viewed from the perspective of score-based generative models \citep{song2020score}, which represent the distribution of data via gradients, and are sampled from using Langevin dynamics \citep{welling2011bayesian}. Diffusion models have been trained by optimizing the variational bound on data likelihood \citep{ho2020denoising}. Modeled as stochastic differential equations \citep{song2020score}, the same training objective has also been shown to enable maximum-likelihood training under specific weighing schemes \citep{song2021maximum}. 
Diffusion models have also demonstrated strong results on conditional image generation tasks. \cite{sohl2015deep} and \cite{song2020score} showed that gradients of a classifier can be used to condition a pre-trained diffusion model. \cite{dhariwal2021diffusion} introduced Classifier-guidance, which achieved state-of-the-art results in image synthesis at the time it was released. \cite{song2020denoising}  introduced Denoising Diffusion Implicit Models (DDIM), greatly accelerating the process of sampling from trained diffusion models as compared to DDPM \citep{ho2020denoising}. \cite{karras2022elucidating} presented a unified view of existing diffusion models from a practical standpoint, enabling develop improved sampling and training techniques to obtain greatly improved results.

\paragraph{Semantic Space of Generative Models.}
The Distributional Hypothesis \citep{harris1954distributional} forms the basis for statistical semantics, characterizing the meaning of linguistic items based on their usage distributions.  This is also closely related to Wittgenstein's use theory of meaning \citep{wittgenstein1953}, often popularized as ``meaning is use". Many methods have been developed in machine learning and Natural Language Processing (NLP) literature to compute these semantic spaces, including Word2Vec
\citep{mikolovDistributedRepresentationsWords2013}. In light of modern Large Language Models (LLMs), \cite{liu2023meaning} defined the space of meanings for autoregressive models to be the distribution over model continuations for any input sequence. This has been used to define notions of semantic distances and semantic containment between textual inputs. \cite{achille2024interpretable} further defined conceptual similarity between images by projecting them into the space of distributions over
complexity-constrained captions, producing similarity scores that strongly correlate with human annotations. \cite{soattoTamingAIBots2023} generalizes these definitions by considering meanings as equivalence classes, where partitions can be induced by either an external agent or the model itself. To evaluate generative distributions, \cite{pillutla2023mauve} quantifies the gap between target and model distributions by estimating the area under divergence frontiers in an embedding space. Vector-based representations such as those obtained from CLIP \cite{radford2021learning}, or in general any sentence embedding model \citep{devlin2018bert,opitz2022sbert} have also been
defined specifically for the computation of semantic distances. Such representations, however, are often difficult to interpret. 

In contrast to these works, \cite{bender2020climbing} argues that training on language alone is insufficient to capture semantics, which they argue requires a notion of ``communicative intents" that are external to language. In this paper, we explore a notion of meaning that is grounded in the distribution of evoked imageries, and present a simple algorithm to compute interpretable distances in this space for the class of text-conditioned diffusion models.

\paragraph{Evaluation and Interpretation of Diffusion Models.}
Common metrics used to evaluate diffusion models are, among many others, the widely-used FID \citep{heusel2017gans} score, Kernel Inception Distance \citep{binkowski2018demystifying}, and the CLIP score \citep{hessel2021clipscore}. Despite these choices, \cite{stein2024exposing} recently discovered that no existing metric used to evaluate diffusion models correlates strongly with human evaluations. While existing evaluations often focus on the quality and diversity of generations, our method is the first to evaluate semantic alignment of the representations learnt by diffusion models.
Several techniques have also been developed to interpret the generation of diffusion models. \cite{kwon2022diffusion,park2023unsupervised} edit the bottleneck representations within the U-Net architecture of diffusion models for semantic image manipulation. \cite{gandikota2023concept} identifies low-rank directions  corresponding to various semantic concepts, and \cite{li2023your} shows that diffusion models can be used for image classification. \cite{kong2023information,kong2023interpretable} frame diffusion models using information theory, improving interpretability of their learnt semantic relations. Orthogonal to these works, our method enables visualizing the semantic relations between textual prompts in natural language learnt by diffusion models via the distributions over their generated imageries.

\section{Method}
We will present a short preliminary on (conditional) diffusion models in \Cref{sec:preliminary}, and derive our algorithm for computing semantic similarity in \Cref{sec:construction}.

\subsection{Preliminary}
\label{sec:preliminary}
Our derivations will leverage \cite{song2020score}'s the SDE formulation of diffusion models when viewed from the lens of score-based generaive modeling. In particular, a (forward) diffusion process $\{ \bm{x}(t)\}_{t=0}^T$ can be modeled as the solution to the following SDE:
\begin{align*}
    d\bm{x} = \bm{f}(\bm{x},t)dt + g(t)d\bm{w}_t
\end{align*}
with drift coefficient $\bm{f}: \mathbb{R}^{d} \times \mathbb{R} \mapsto \mathbb{R}^{d}$ and (scalar) diffusion coefficient $g : \mathbb{R} \mapsto \mathbb{R}$, where $\bm{w}_t$ is standard Brownian motion. We constrain the timesteps $t$ such that $t \in [0,T]$, where $\bm{x}(0)$ represents the distribution of ``fully-denoised" images generated by the diffusion model. We also assume that by construction, the prior at time $T$ is known and distributed according to $\bm{x}(T) \sim \pi$, where often $\pi = N(0,I)$. 

Once trained, we can view text-conditioned diffusion models as a map $s_\theta(\bm{x},t|y)$ parameterized by $\theta$ and conditioned on a textual prompt $y \in \mathcal{Y}$ that is used to approximate the score function \citep{song2020score}, where $s_\theta(\cdot,t|y) : \mathbb{R}^d \mapsto \mathbb{R}^d$ and $\mathcal{Y}$ is the set of textual expressions. As such, each conditional model $s_\theta(\bm{x},t|y)$ defines a reverse-time SDE given by:
\begin{align}
\label{eqn:sde}
    d\bm{x} = [f(\bm{x}, t) - g(t)^2 s_\theta(\bm{x},t|y)] dt + g(t) d \bar{\bm{w}}_t
\end{align}
where $\bar{\bm{w}}_t$ is the Brownian motion running backwards in time from $t=T$ to $t=0$. For ease of notation, we will denote $\mu_\theta(\bm{x},t,y) := [f(\bm{x}, t) - g(t)^2 s_\theta(\bm{x},t|y)]$.

\subsection{Construction}
\label{sec:construction}

\begin{algorithm}[t]
\caption{Conjuring Semantic Similarity}\label{alg:similarity}
\begin{algorithmic}
\Require Diffusion model $s_\theta$, Prompts $y_1, y_2$, Monte-Carlo steps $k$
\State Initialize $d = 0$
\For{$i = 1 \ldots k$}
\State $x_T \gets$ Sample from initial distribution $\pi$
\State $\hat{x}_{T}, \ldots, \hat{x}_0 \gets $ Denoise $x_T$ conditioned on $y_1$
\State $\tilde{x}_{T}, \ldots, \tilde{x}_0 \gets $ Denoise $x_T$ conditioned on $y_2$
\State $d \gets d + \frac{1}{T} \sum_{t=1}^{T} \left\Vert s_\theta(\hat{x}_t, t|y_1) - s_\theta(\hat{x}_t, t|y_2)\right\Vert_2^2$
\State $d \gets d + \frac{1}{T} \sum_{t=1}^{T} \left\Vert s_\theta(\tilde{x}_t, t|y_1) - s_\theta(\tilde{x}_t, t|y_2)\right\Vert_2^2$
\EndFor
\State\Return{$d / k$} \Comment{Return similarity score}
\end{algorithmic}
\end{algorithm}

Given two textual prompts $y_1$ and $y_2$, we obtain two separate diffusion SDEs in the space of images using \cref{eqn:sde}, which are given by 
\begin{align}
    &d\bm{x}_1 = \mu_\theta(\bm{x}_1,t,y_1) dt + g(t) d \bar{\bm{w}}_t  \label{eqn:sde-y1} \\
    &d\bm{x}_2 = \mu_\theta(\bm{x}_2,t,y_2) dt + g(t) d \bar{\bm{w}}_t \label{eqn:sde-y2}
\end{align}

We assume the standard conditions for existence and uniqueness of their solutions, in particular for all $t \in [0,T]$ we have $\mathbb{E} \left[ \Vert \bm{x}(t) \Vert_2^2 \right] \leq \infty$, and for all $y \in \mathcal{Y}$ and $x,x' \in \mathbb{R}^d$, there exists constants $C, D$ such that $\Vert \mu_\theta(x,t,y) \Vert_2 + \Vert g(t) \Vert_2 \leq C(1 + \Vert x \Vert_2)$ and $\Vert \mu_\theta(x,t,y) - \mu_\theta(x',t,y) \Vert_2 \leq D \Vert x-x' \Vert_2$. We further assume Novikov's Condition holds for all pairs $y_1, y_2 \in \mathcal{Y}$, in particular we are guaranteed the following: $\mathbb{E}\left[\exp \left({\frac{1}{2} \int_0^T \Vert \mu_\theta(\bm{x}, t, y_2) - \mu_\theta(\bm{x}, t, y_1)\Vert_2^2 dt } \right)\right] < \infty$.

Since our goal is to define a semantic distance between textual prompts $y_1$ and $y_2$ by comparing the distributions over images that they conjure, we can achieve this by computing a discrepancy function between the respective SDEs that they induce. In particular, we will use the Jeffreys divergence, which is simply the symmetrized Kullback–Leibler (KL) divergence between two SDEs. In the following, we show how this divergence can be computed via a Monte-Carlo approach.

Denote the path measures associated with \cref{eqn:sde-y1} and \cref{eqn:sde-y2} respectively to be $\mathbb{P}_1$ and $\mathbb{P}_2$. Then, the KL divergence between the two SDEs are defined via
\begin{align*}
    D_{KL}(\mathbb{P}_2 || \mathbb{P}_1) = -\mathbb{E}_{\mathbb{P}_2} \log \left( \frac{d \mathbb{P}_1}{d \mathbb{P}_2} \right)
\end{align*}
where $\frac{d \mathbb{P}_1}{d \mathbb{P}_2}$ is the Radon-Nikodym derivative.
By Girsanov theorem \citep{girsanov1960transforming}, we can compute this derivative as the stochastic exponential given by
\begin{align*}
  \exp{ \left( \int_0^T -\frac{1}{g(t)} \left(\mu_\theta(\bm{x}, t, y_1) - \mu_\theta(\bm{x}, t, y_2) \right) d\bar{\bm{w}}_t - \frac{1}{2} \int_0^T \frac{1}{g(t)^2} \Vert \mu_\theta(\bm{x}, t, y_1) - \mu_\theta(\bm{x}, t,y_2)\Vert_2^2 dt \right) }
\end{align*}

Next, paralleling the derivation in \cite{song2021maximum}, Novikov's Condition guarantees that the stochastic integrand term $-\frac{1}{g(t)} \left(\mu_\theta(\bm{x}, t, y_1) - \mu_\theta(\bm{x}, t, y_2) \right)$ is a Martingale. As a result, taking expectation under $\mathbb{P}_2$ would reduce it to zero, leaving only the drift term $- \frac{1}{2} \int_0^T \frac{1}{g(t)^2} \Vert \mu_\theta(\bm{x}, t, y_1) - \mu_\theta(\bm{x}, t,y_2)\Vert_2^2 dt$. Consequently, the KL divergence between the two SDEs can be simplified as
\begin{align*}
    D_{KL}(\mathbb{P}_2 || \mathbb{P}_1) &= \frac{1}{2} \mathbb{E}_{\mathbb{P}_2} \left[ \int_0^T \frac{1}{g(t)^2} \Vert \mu_\theta(\bm{x}, t, y_1) - \mu_\theta(\bm{x}, t, y_2)\Vert_2^2 dt \right] \\
    &= \frac{1}{2} \mathbb{E}_{\mathbb{P}_2} \left[ \int_0^T g(t)^2 \Vert s_\theta(\bm{x}, t| y_1) - s_\theta(\bm{x}, t| y_2)\Vert_2^2 dt \right]
\end{align*}
We can symmetrize to form our desired distance function:
\begin{align*}
    d_{ours}(y_1, y_2) := D_{KL}(\mathbb{P}_2 || \mathbb{P}_1) + D_{KL}(\mathbb{P}_1 || \mathbb{P}_2)
\end{align*}
which, ignoring constants, can be written as
\begin{align*}
d_{ours}(y_1, y_2) &= \mathbb{E}_{t \sim \text{unif}([0,T]), \bm{x} \sim \frac{1}{2} p_t(\bm{x}|y_1) + \frac{1}{2} p_t(\bm{x}|y_2)} \left[ g(t)^2 \Vert s_\theta(\bm{x}, t|y_1) - s_\theta(\bm{x}, t|y_2)\Vert_2^2  \right]
\end{align*}
where $p_t(\cdot|y)$ is the distribution of noisy images at timestep $t$. Similar to how losses at different timesteps are weighted uniformly in the training of real-world diffusion models (\textit{e.g.} $L_{simple}$ proposed by \cite{ho2020denoising}), we set $g(t)$ to be constant, in particular $g(t) = 1$, to simplify our algorithm such that it does not have to be tailored specifically to each choice of scheduler.

We can compute this resulting semantic distance using Monte-Carlo by discretizing the timesteps to $t \in \{1, \ldots, T\}$. In particular, this is computed in practice via sampling an initial noise vector $\bm{x}(T) \sim \pi$, and denoising it with both $y_1$ and $y_2$ to obtain a sequence of samples $x_t$'s, and computing the difference in predictions $\Vert s_\theta(x_t, t|y_1) - s_\theta(x_t, t|y_2)\Vert_2^2$ at each denoising timestep. We describe this process in Algorithm \ref{alg:similarity}.

\section{Experiments}
We describe implementation details in \Cref{sec:implementation}, empirical validation for our definitions in \Cref{sec:results}, and ablations in \Cref{sec:ablations}.

\subsection{Implementation Details}
\label{sec:implementation}
We use Stable Diffusion v1.4 \citep{rombach2022high}, a text-conditioned diffusion model, for all our experiments. For sampling, we use classifier-free guidance \citep{ho2022classifier} with guidance scale of $7.5$, and sample using the LMS Scheduler \citep{karras2022elucidating}. We specify image sizes to be $512 \times 512$, but note that Stable Diffusion v1.4 uses latent diffusion, as such model predictions are in practice of dimension $64 \times 64$. We compute the Euclidean distance directly in this space. However, for visualization experiments such as in \Cref{fig:conjuring-semantic-similarity}, we decode the predictions using the VAE before plotting them. We set $T=10$ in our experiments to ensure computational feasibility, after ablating over other choices in \Cref{tab:choice-of-T}. We perform all our experiments with a single RTX 4090 GPU, and each Monte-Carlo step takes around 2.0s to complete in our naive implementation.

As a technicality, while we model the denoising direction term specified in the reverse-time SDE in \cref{eqn:sde} as the (text-conditioned) model output $s_\theta(\cdot|y)$, the exact implementation varies depending on the text-conditioning method used. However, we note that in the case of classifier-free guidance, the resulting distance computed using the model output as $s_\theta(\cdot|y)$ is equivalent to that computed using classifier-guidance directions up to proportionality.

\subsection{Baselines}
\label{sec:baselines}

While there exists no comparable baselines for quantifying semantic similarity in text-conditioned diffusion models at the time of writing, we present several derivatives of our method below which are directly comparable against:

\textbf{Prediction at initial timestep:} We compare the one-step predicted noise vector at the initial denoising timestep. This is defined as $d_{\text{initial}} := \mathbb{E}_{\bf{x} \sim \pi} \left[  \Vert s_\theta(\bm{x}, T|y_1) - s_\theta(\bm{x}, T|y_2)\Vert_2^2  \right]$.

\textbf{Prediction at final timestep:} We compare the predicted noise vector at the final denoising timestep. This is defined as $d_{\text{final}} := \mathbb{E}_{ \bm{x} \sim \frac{1}{2} p_0(\bm{x}|y_1) + \frac{1}{2} p_0(\bm{x}|y_2)} \left[  \Vert s_\theta(\bm{x}, 0|y_1) - s_\theta(\bm{x}, 0|y_2)\Vert_2^2  \right]$.

\textbf{Direct output comparisons:} For the same initial condition, we directly compute the difference between the images produced by the two different labels. This is defined as $d_{\text{output}} := \mathbb{E}_{ \bm{x} \sim \pi} \left[  \Vert \psi_\theta(\bm{x}|y_1) - \psi_\theta(\bm{x}, |y_2)\Vert_2^2  \right]$, where $\psi_\theta(\cdot|y) : \mathbb{R}^d \mapsto \mathbb{R}^d$ represents the full reverse diffusion process from the noise prior $\pi$ at time $T$ to the output distribution at time $0$.

\textbf{KL-Divergence:} We evaluate the non-symmetrized version of our method, computed via the KL divergence between the SDEs obtained from different prompts: $d_{\text{ours-KL}}(y_1, y_2) := \mathbb{E}_{t \sim \text{unif}([0,T]), \bm{x} \sim p_t(\bm{x}|y_1)} \left[ g(t)^2 \Vert s_\theta(\bm{x}, t|y_1) - s_\theta(\bm{x}, t|y_2)\Vert_2^2  \right]$

Using the same parameters as our proposed method, we implement all baselines via Monte-Carlo sampling.

\subsection{Empirical Validation}
\label{sec:results}

\begin{table}[t!]
    \footnotesize
    \centering
    \setlength{\tabcolsep}{2.5pt}
    \caption{Comparison with zero-shot methods on Semantic Textual Similarity benchmarks, evaluated via Spearman Correlation (higher is better). $\ast;\dagger;\ddagger$ indicate results taken from \cite{ni2021sentence,gao2021simcse,liu2023meaning} respectively. Expectedly, our zero-shot approach does not perform as well as embedding models such as CLIP \citep{radford2021learning} and SimCSE-BERT \citep{gao2021simcse}, which are trained specifically for semantic comparison tasks. Nevertheless, semantic structures extracted from text-conditioned diffusion models (StableDiffusion) using our method are still well-aligned with human annotators, rivaling those extracted from autoregressive Large Language Models while outperforming encoder-based language models such as BERT \citep{devlin2018bert}.}
    \resizebox{\textwidth}{!}{%
    \begin{tabular}{lrrrrrrrr}\toprule
 &STS-B &STS12 &STS13 &STS14 &STS15 &STS16 & SICK-R & Avg \\
\midrule
\multicolumn{7}{l}{\textit{Contrastive-Trained Embedding Models}} \\
\midrule
CLIP-ViTL14$^\ddagger$ \citep{radford2021learning} &65.5 &67.7 &68.5 &58.0 &67.1 &73.6 & 68.6 & 67.0 $\pm$ 4.3  \\
IS-BERT$^\dagger$ \citep{zhang2020unsupervised} & 56.8 & 69.2 & 61.2 & 75.2 & 70.2 & 69.2 & 64.3 & 66.6 $\pm$ 5.7 \\
SimCSE-BERT$^\dagger$ \citep{gao2021simcse} & 68.4 & 82.4 & 74.4 & 80.9 & 78.6 & 76.9 &  72.2 & \textbf{76.3} $\pm$ 4.6 \\
\midrule
\multicolumn{7}{l}{\textit{Zero-Shot Encoder-based Models}} \\
\midrule
BERT-CLS$^\ast$ \citep{devlin2018bert} &16.5 &20.2 &30.0 &20.1 &36.9 &38.1 & 42.6 & 29.2 $\pm$ 9.6 \\
BERT-mean$^\ast$ \citep{devlin2018bert} &45.4 &38.8 &58.0 &58.0 &63.1 &61.1 & 58.4 & 54.8 $\pm$ 8.3 \\
BERT Large-mean$^\ast$ \citep{devlin2018bert} & 47.0 & 27.7 & 55.8 & 44.5 & 51.7 & 61.9 & 53.9 & 48.9 $\pm$ 10.2 \\
RoBERTa Large-mean$^\ast$ \citep{liu2019roberta} & 50.6 & 33.6 & 57.2 & 45.7 & 63.0 & 61.2 & 58.4 & 52.8 $\pm$ 9.6 \\
ST5-Enc-mean (Large)$^\ast$ \citep{ni2021sentence} &56.3 &28.0 &52.6 &41.4 &61.3 &63.6 & 59.5 & 51.8 $\pm$ 11.9 \\
ST5-Enc-mean (11B)$^\ast$ \citep{ni2021sentence} &62.8 &35.0 &60.2 &47.6 &66.4 &70.6 & 63.6 & \textbf{58.0} $\pm$ 11.5 \\
\midrule
\multicolumn{7}{l}{\textit{Autoregressive Models (Meanings as Trajectories \citep{liu2023meaning})}}  \\
\midrule
GPT-2$^\ddagger$ &  55.2 & 39.9 & 42.6 & 30.5 & 52.4 & 62.7 & 62.0 & 49.3 $\pm$ 11.2 \\
GPT-2-XL$^\ddagger$ & 62.1 & 43.6 & 54.8 & 37.7 & 61.3 & 68.2 & 68.4 & 56.5 $\pm$ 11.1 \\
Falcon-7B$^\ddagger$ & 67.7 & 56.3 & 66.5 & 53.0 & 67.4 & 75.5 & 73.5 & 65.7 $\pm$ 7.7 \\
LLaMA-13B$^\ddagger$ & 70.6 & 52.5 & 65.9 & 53.2 & 67.8 & 74.1 & 73.0 & 65.3 $\pm$ 8.3 \\
LLaMA-33B$^\ddagger$ & 71.5 & 52.5 & 70.6 & 54.6 & 69.1 & 75.2 & 73.0 & \textbf{66.6} $\pm$ 8.5 \\
\midrule
\multicolumn{7}{l}{\textit{Text-Conditioned Diffusion Models (StableDiffusion)}} \\
\midrule
{Initial Timestep Prediction} &  55.8 &  46.7 &  53.4 &  47.2 &  54.3 &  57.9 &  56.0 &  53.0 $\pm$ 4.1 \\
{Final Timestep Prediction} & 64.9 &  41.6 &  56.4 & 51.0 & 65.2 & 60.2 & 58.9 & 56.9 $\pm$ 7.7 \\
{Direct Output Comparison} &  57.0 &  44.7 & 45.6 &  43.3 &  58.5 &  56.2 &  53.5 &  51.3 $\pm$ 6.0 \\
 Conjuring Semantic Similarity (KL-Div) & 69.1 & 56.9 & 60.6 & 59.5 &  71.5 &  65.7 &  64.8 &  64.0 $\pm$ 4.9 \\
Conjuring Semantic Similarity & 70.3 & 57.9 & 61.0 & 60.8 & 73.6 & 67.9 & 66.0 & \textbf{65.4} $\pm$ 5.3\\
\bottomrule
    \end{tabular}
    }
    \label{tab:stsb-1}
\vspace{-1em}
\end{table}

While our work defines a notion of semantic distance grounded in evoked imagery, the validity of this definition hinges on its use as a measure of similarity that aligns with humans'. In particular, we should expect our definition to produce measurements of similarity that agree often with human annotators (which can be viewed as ``ground-truth").

To quantify this, we use the Semantic Textual Similarity (STS) \citep{agirre2012semeval,agirre2013sem,agirre2014semeval,agirre2015semeval,agirre2016semeval,cer2017semeval} and Sentences Involving Compositional Knowledge (SICK-R) \citep{marelli2014semeval} datasets, containing pairs of sentences each labelled by human annotators with a semantic similarity score ranging from 0-5. We then use our method to compute the image-grounded similarity score, and measure their resulting Spearman Correlation with the annotations.

Interestingly, our experiments in Table \ref{tab:stsb-1} show that our visually-grounded similarity scores exhibit significant degrees of correlation with that annotated by humans. While, expectedly, our method presently lags behind embedding models trained specifically for semantic comparison tasks, we show that our visually-grounded similarity scores can rival that produced by existing large language models up to 33B in size. Since our work is the first to formalize and evaluate semantic alignment for this class of conditioned diffusion models, we have also included results from baselines with which our approach is directly comparable against. Our method convincingly outperforms all baseline methods for evaluating semantic similarity in diffusion models. 

We further remark that in most existing text-conditioned diffusion models, the representation structures that can be captured by our method are limited by those learnt by text-encoder models such as CLIP \citep{radford2021learning}, since these encoders are often used to pre-process textual prompts. In light of this, the experiments also suggest that our method can be an effective metric to quantify how well representation structures learnt by these text-encoders have been distilled to the resulting diffusion model. We explore this in the following section, where we study how faithfully learnt semantic relations between words transfer from text-encoders to the full diffusion model.

\begin{figure}[h]
    \centering
    \includegraphics[width=0.49\linewidth]{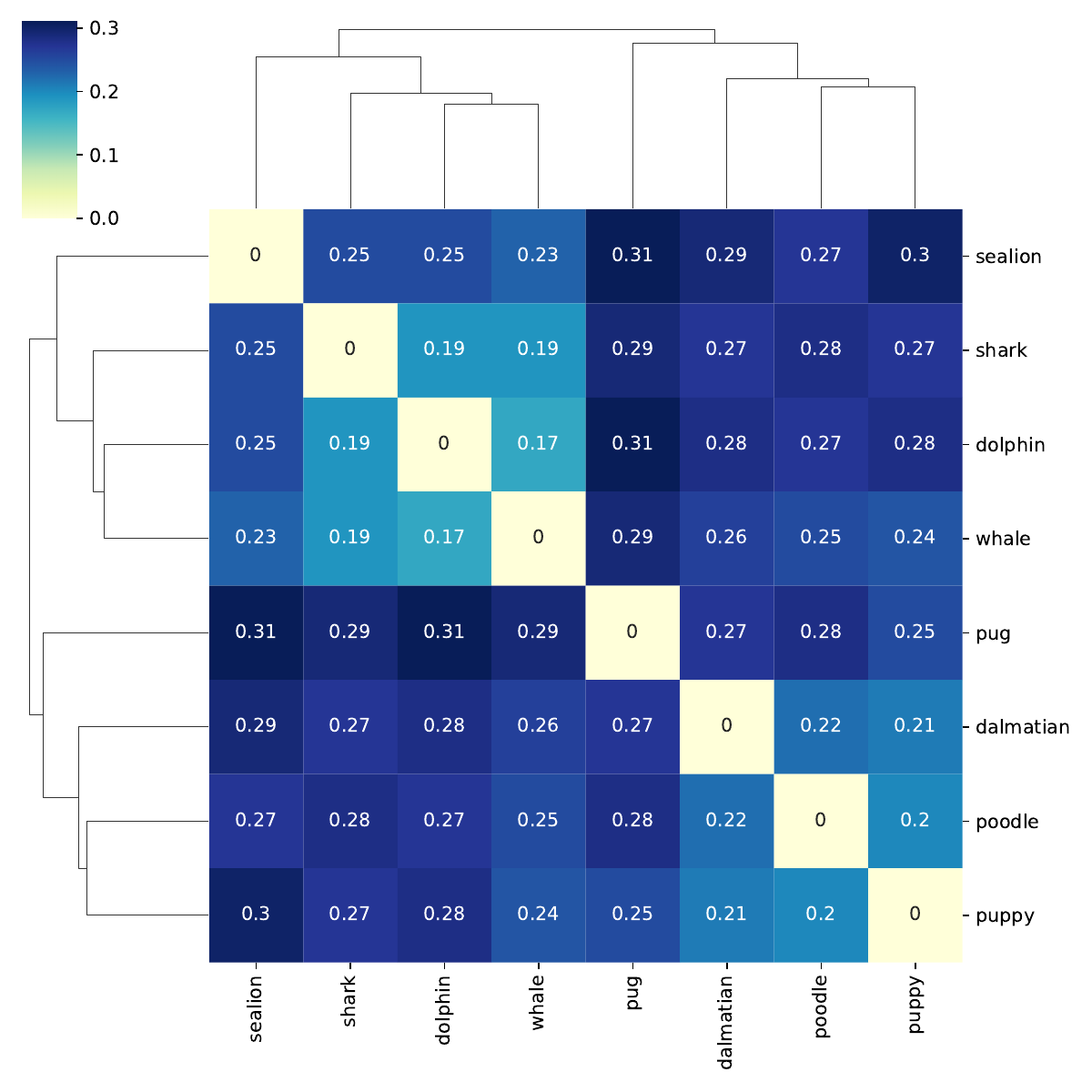}
    \hfill
    \includegraphics[width=0.49\linewidth]{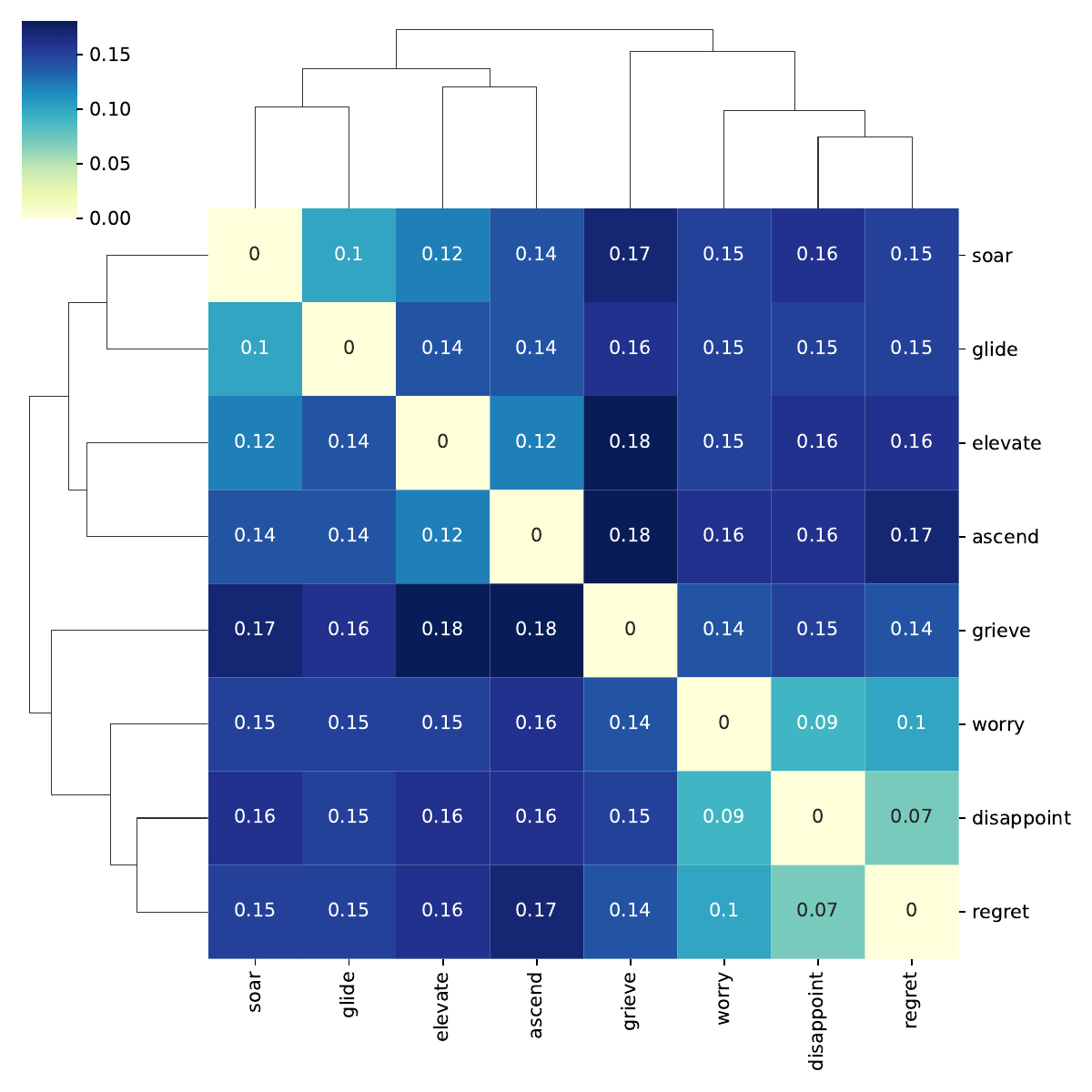}
    \caption{Qualitative evaluation of conjured semantic similarity, visualized using pairwise distance matrices and hierarchial clustering. \textbf{(Left)} shows that nouns cluster based on shared hypernym classes: Dogs (puppy, poodle, dalmatian, pug) form a visible cluster in the bottom-right 4x4 block, while marine animals (whale, shark, dolphin, sealion) form another cluster in the top-left 4x4 block. \textbf{(Right)} shows that the same pattern holds for flying-related action verbs (elevate, ascend, soar, glide) v.s. negative stative verbs (disappoint, grieve, worry, regret).}
    \label{fig:clustering-of-words}
\end{figure}

In \Cref{fig:clustering-of-words}, we also qualitatively evaluate our method on measuring semantic similarity between various words. From the resulting pairwise similarity matrices, we can observe that words are closer in terms of their common hypernym class tend to cluster together, showing that our method can effectively capture word taxonomies. For instance, nouns that describe types of dogs cluster together, and nouns that describe types of marine animals similarly
form another cluster. These two clusters are separate, in the sense that distances of words across these clusters are large than those within each cluster.

\subsection{Empirical Analysis}
\label{sec:ablations}
In this section, we run ablation studies using the STS-B dataset as a benchmark to explore the design space of our method, and improve its computational efficiency. We further analyze the failure modes of diffusion models through comparing semantic alignment of words belonging to different parts of speech.

\begin{figure}[h]
    \centering
    \includegraphics[width=0.52\textwidth]{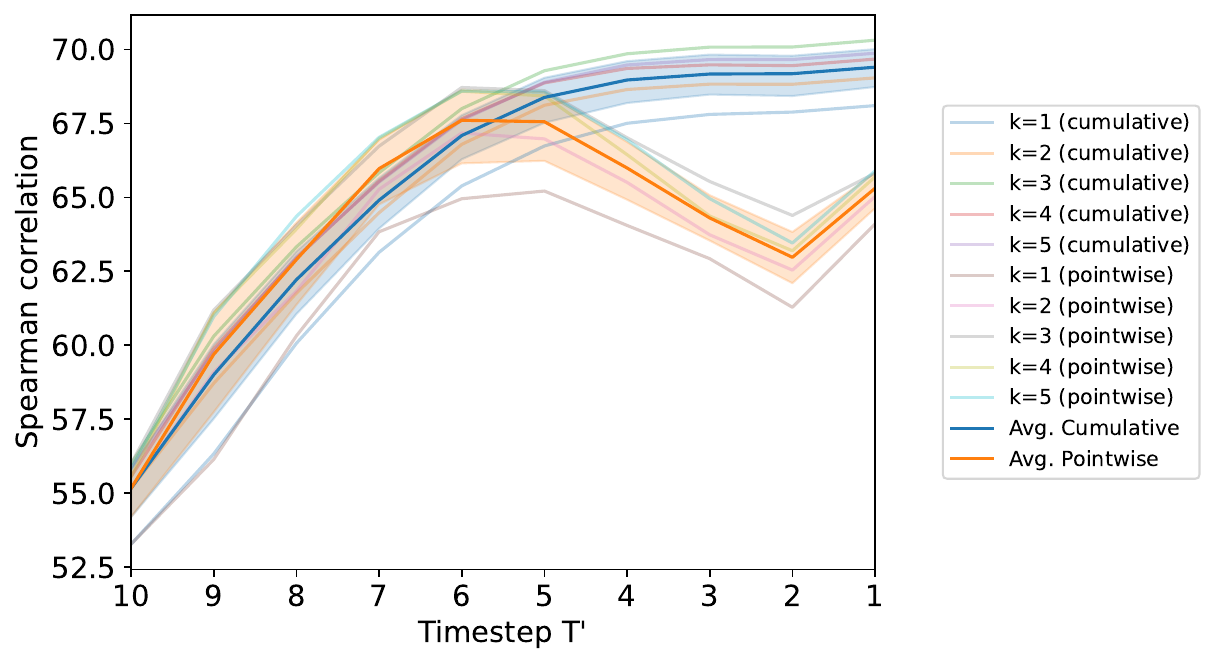}
    \hfill
    \includegraphics[width=0.39\textwidth]{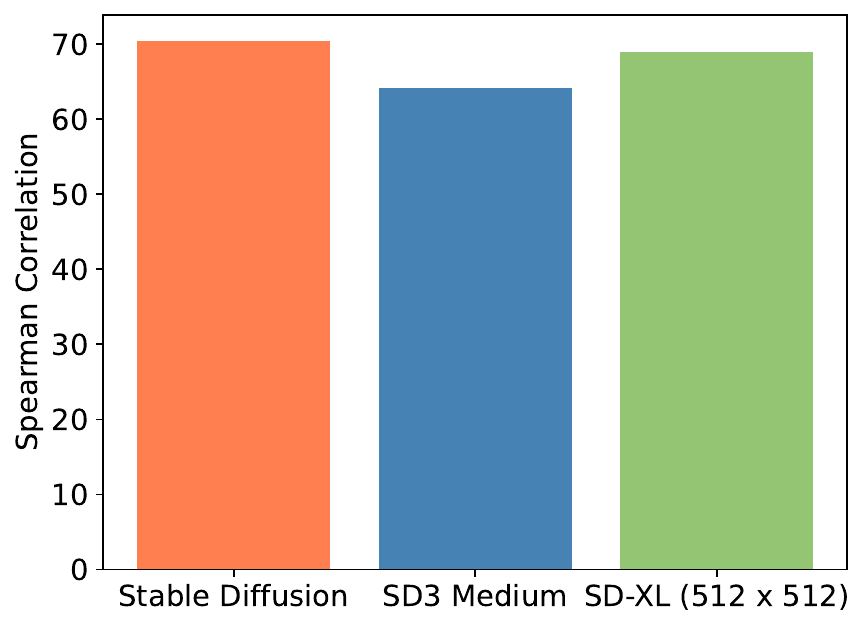}
    \caption{\textbf{(Left:)} We ablate over different choices of priors over timesteps -- a uniform distribution over timesteps $\{T', \ldots, T\}$ where $T' \leq T=10$, represented by the blue line (cumulative), and the Direc Delta on any particular timestep $T' \in \{1, \ldots, T\}$, represented by the orange line (pointwise). We show that a uniform prior over all timesteps gives the best results. The same plot also ablates over the number of Monte-Carlo samples, $k \in \{1, \ldots, 5\}$, where we conclude that only few iterations are required for convergence. \textbf{(Right:)} We further ablate over different choices of diffusion models, and show that results remain relatively consistent across the tested choices.}
    \label{fig:cumulative-vs-fixed-timesteps}
\end{figure}

\paragraph{Prior over timestep distribution.}
In our algorithm, we placed a uniform prior over timesteps $\{1, \ldots, T\}$. In \Cref{fig:cumulative-vs-fixed-timesteps}, we show that this simple choice works best when evaluated on alignment with human annotators on STS-B, as compared to other choices such as considering only a uniform prior over the subset $\{T', \ldots, T\}$ (cumulative from $T'$ to $T$) where $T' \leq T$, or a Direc delta on any particular timestep $T' \in \{1, \ldots, T\}$ (pointwise).

\paragraph{Number of Monte-Carlo Steps.} The computational feasibility of our method depends on the number of Monte-Carlo steps (\textit{i.e.} $k$ in \Cref{alg:similarity}) required to produce a reliable approximate of the desired distance. We ablate over choices of $k \in \{1, \ldots, 5\}$ in \Cref{fig:cumulative-vs-fixed-timesteps} and show that the deviation of scores when evaluated on the STS-B dataset is small ($\pm 0.77$) across different choices of $k$. This finding is promising, as it implies that our method can be computationally efficient, requiring only a small number of Monte-Carlo iterations to converge.

\paragraph{Choice of Stable Diffusion Model.} On the right of \Cref{fig:cumulative-vs-fixed-timesteps}, we also show that our results are relatively consistent across several versions of Stable Diffusion models, including Stable Diffusion XL and Stable Diffusion 3 Medium.

\paragraph{Error Analysis.} Here, we evaluate our method on word similarity datasets, categorized by the part of speech (POS) from which the words originate. We use RG65 \citep{rubenstein1965contextual} and SimLex-999 \citep{hill2015simlex} for our evaluation, which contains pairs of words and their semantic similarity score as annotated by humans. The former consists of nouns, while the latter includes adjectives, nouns, and verbs. Our analysis in \Cref{tab:simlex} compares the representations learnt by the text encoder of the diffusion model to that of the model itself (extracted via our approach). Interestingly, we observe that while the semantic relations between nouns are largely preserved, the semantic relations between verbs and adjectives tend to deteriorate after learning the reverse diffusion process for image generation.

\begin{table}[h]
    \centering
    
    \caption{ We evaluate the spearman correlation between semantic similarity scores obtained via our method, and that labeled by human annotators. We use the text encoder bottleneck of the StableDiffusion model as the paragon. While the semantic properties of nouns are largely preserved from learning the diffusion process, this comes at the expense of semantics of adjectives and verbs.}
    \begin{tabular}{lcccccc}
    \toprule
         & RG65  & SimLex & SimLex (Adj) & SimLex (Noun) & SimLex (Verb) \\
    \midrule
         Paragon & 77.5  & 34.2 & 40.6 & 42.8 & 10.3 \\
         StableDiffusion & 60.2  &  20.5 & 34.0 & 30.0 & -14.3 \\
     \bottomrule
    \end{tabular}
    \label{tab:simlex}
\end{table}

\begin{table}[th]
    \centering
    \caption{Ablation on choices for $T$ on the STS-B dataset: We show that variance with respect to the choice of $T$ is small, allowing semantic distances to be computed efficiently by using lower values.}
    \begin{tabular}{rcccccc}
    \toprule
     & $T=5$ & $T=10$ & $T=15$ & $T=20$ & $T=30$ & $T=50$ \\
     \midrule
     Spearman Corr. & $70.1$ & $70.3$ & $70.1$ & $68.9$ & $70.2$ & $69.5$ \\
    \bottomrule
    \end{tabular}
    \label{tab:choice-of-T}
\end{table}

\section{Discussion and Limitations}

Our method has several limitations. Our work does not, nor is it meant to, provide a superior, general-purpose metric for computing semantic similarity (\textit{e.g.} for applications such as retrieval). Indeed, imageries might indeed not be sufficient to fully capture the meaning of certain expressions, such as mathematical abstractions (like `imaginary numbers') and metaphysical concepts (like `conscience').
Furthermore, many modern diffusion models use a pre-trained text-encoder to pre-process textual prompts. This means that representation structures obtained from the diffusion model outputs would be bottle-necked by those learnt by the text encoder. 
The meanings of textual expressions can often be ambiguous or open to multiple (subjective) interpretations. Our method does not resolve this linguistic ambiguity, since generated images are conditioned on the same potentially ambiguous text they are meant to clarify. Instead, it simply reveals which visual interpretation the model has associated with that expression.
 
 Finally, computation costs also remain a key limitation of our method, since it requires several inference passes through the diffusion model to compute a single semantic similarity score, mitigated only partially by the conclusions of our ablation study on the required number of iterations.

Nevertheless, our method is the first to show that textual representations can be meaningfully compared for diffusion models by ``grounding" them in the space of conjured images.
 We introduce the first method for evaluating semantic alignment of text-conditioned diffusion models. In particular, our work enables not only qualifying (via visual `explanations'), but also quantifying the alignment of this resulting semantic space with that of human annotators. Our method also enables fine-grained analysis of the failure modes of existing diffusion models, pinpointing specific areas where they align poorly with human annotators. Our general framework of conjuring semantic similarity is applicable to the broader class of image generative models, and we leave the exploration of applications beyond diffusion models to future work. 

\section{Acknowledgements}
Research supported by ONR N00014-19-1-2229.

\bibliography{iclr2026_conference}
\bibliographystyle{iclr2026_conference}
\clearpage
\appendix
\section{Additional Visualizations}

\begin{figure}[h]
    \centering
    \includegraphics[width=\linewidth]{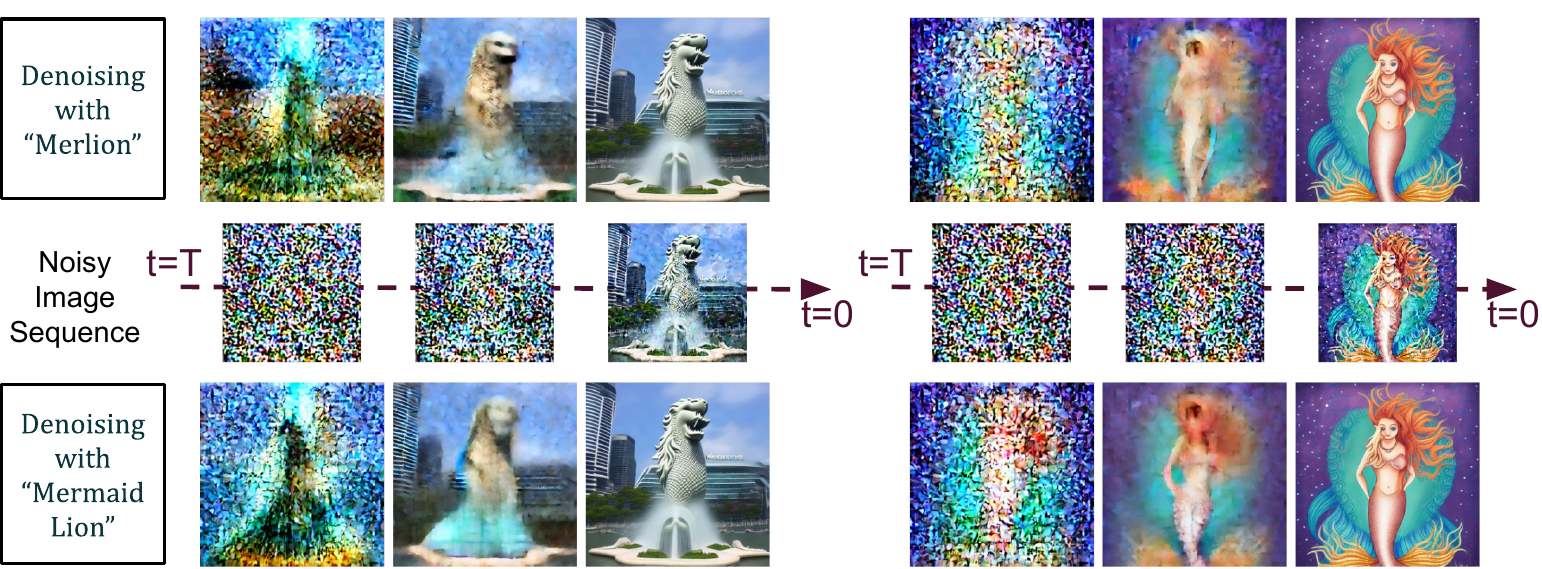}
    \caption{``Merlion" vs ``Mermaid Lion": While both prompts express compositions of the same set of objects, the model associates different meanings with ``Merlion" as opposed to ``Mermaid + Lion", where the former is associated to the mascot of Singapore, while the latter is a mermaid with hair resembling a lion's mane.}
    \label{fig:merlion}
\end{figure}

\begin{figure}[h]
    \centering
    \includegraphics[width=\linewidth]{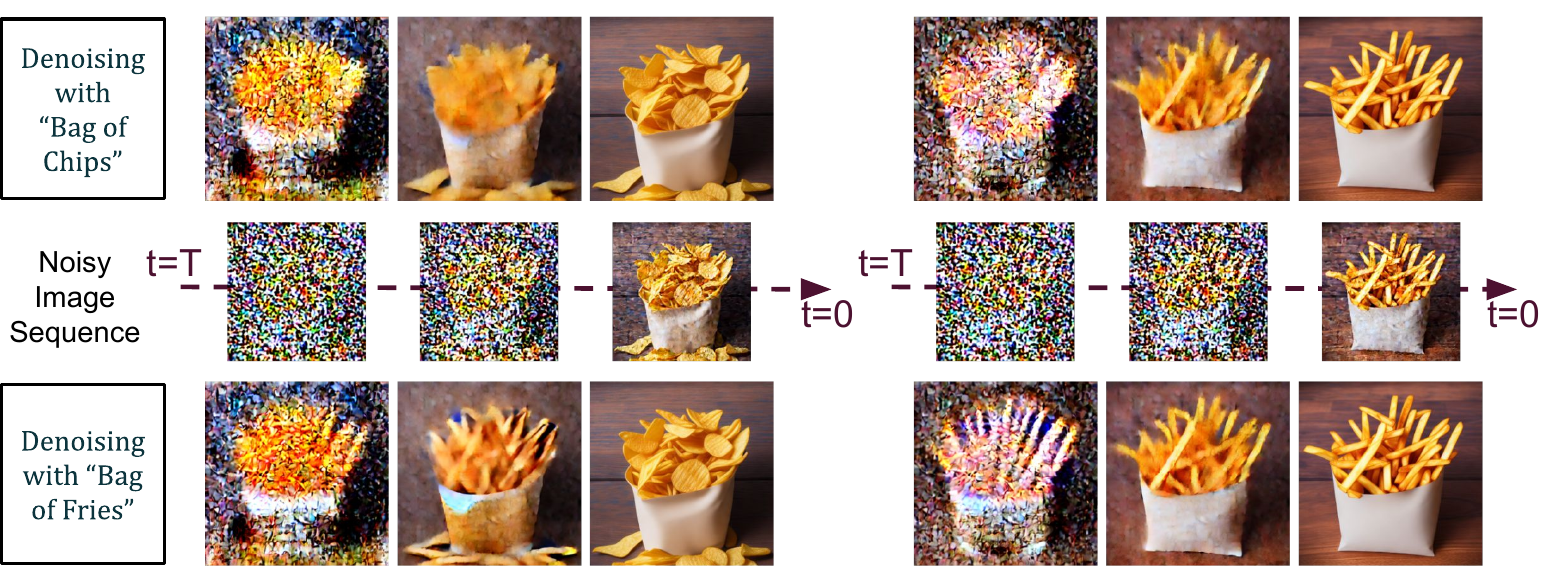}
    \caption{``Bag of Chips" vs ``Bag of Fries": The interpretation of ``chips" depends on cultural background (US vs UK), but the interpretation of ``fries" is relatively non-ambiguous. Interestingly, this observation can be visualized when computing semantic similarity with our method. We see that on the left of the figure (second image column), the model attempts to convert a picture of chips (US) into fries by changing the rounded textures into sharper rectangular ones, when denoised with ``Bag of Fries". On the other hand, pictures of fries still remain relatively identifiable as fries (fifth image column) when denoised using ``Bag of Chips". }
    \label{fig:chips}
\end{figure}

\section{Additional Experiment Details}
We detail the datasets used below:

\emph{Semantic Textual Similarity (STS) \citep{agirre2012semeval,agirre2013sem,agirre2014semeval,agirre2015semeval,agirre2016semeval,cer2017semeval}}: Each of the STS datasets provides pairs of sentences, and a human-labeled score between 0-5 based on how semantic similar each pair is. We use the validation splits in STS-B, containing 1.5K example pairs, and the test splits in STS-12 to STS-16, containing 3.11K, 1,5K, 3.75K, 3K, and 1.19K example pairs respectively. We use the Spearman coefficient (scaled by 100$\times$) to evaluate correlation with these human-annotated similarity scores. 

\emph{Sentences Involving Compositional Knowledge (SICK-R) \citep{marelli2014semeval}}: The SICK-R dataset contains 9.93K pairs of examples which involve compositional knowledge, and similarly provides a human-labeled semantic similarity score for each pair. Below contains an example:
\begin{description}[
    font=\normalfont\bfseries, %
    align=right,              %
    labelwidth=2.5cm,         %
    leftmargin=\dimexpr\labelwidth+\labelsep %
]
    \item[Sentence A:] A man with a jersey is dunking the ball at a basketball game
    \item[Sentence B:] The ball is being dunked by a man with a jersey at a basketball game
    \item[Relatedness Score:] 4.9\\
\end{description}

\emph{RG65\citep{rubenstein1965contextual}} contains human similarity ratings for 65 English noun pairs. 

\emph{SimLex-999\citep{hill2015simlex}} contains 999 different concrete and abstract pairs of adjectives, nouns, and verbs. Each pair is also scored by annotators based on their semantic similarity.

\end{document}